\documentclass[10pt,twocolumn,letterpaper]{article}

\usepackage{cvpr}              
\newtheorem{theorem}{Theorem}
\newtheorem{lemma}[theorem]{Lemma}
\usepackage{amsmath}
\usepackage{epsfig}
\usepackage{amsfonts}
\usepackage{graphicx}
\usepackage{amsmath}
\usepackage{colortbl}
\usepackage{amssymb}
\usepackage{booktabs}
\usepackage{enumerate}
\usepackage{stfloats}
\usepackage{algorithm} %
\usepackage{algpseudocode}

\usepackage{multirow}
\usepackage{graphicx}
\usepackage{color}
\usepackage{orcidlink}
\usepackage{hyperref}
\hypersetup{
    colorlinks=true,
    linkcolor=blue,
    filecolor=blue,      
    urlcolor=blue,
    citecolor=green,
}

\usepackage[english]{babel}

\def\paperID{7843} 
\def\confName{CVPR}
\def\confYear{2025}

\title{Loss-Minimizing Model Compression via Joint Factorization Optimization }

\author{Boyang Zhang$^{1,2}$, Daning Cheng$^1$, Yunquan Zhang$^{1,2}$, Fangming Liu$^2$, Jiake Tian$^3$ \\
$^1$Institute of Computing Technology, Chinese Academy of Sciences, Beijing, China \\
University of Chinese Academy of Sciences, Beijing, China\\
$^2$Peng Cheng Laboratory, Shenzhen, China\\
$^3$the School of Microelectronics, South China University of Technology, Guangzhou, China\\
{\tt\small zhangby01@pcl.ac.cn; chengdaning@ict.ac.cn; zyq@ict.ac.cn;}\\
{\tt\small fangminghk@gmail.com; mijiake@mail.scut.edu.cn}
}

\begin{document}
\maketitle
\begin{abstract}
Low-rank factorization is a popular model compression technique that minimizes the error $\delta$ between approximated and original weight matrices. Despite achieving performances close to the original models when $\delta$ is optimized, a performance discrepancy remains due to the separate optimization processes for low-rank factorization and model performance, resulting in unavoidable losses. We address this issue by introducing a novel joint optimization strategy for lossless low-rank weight factorization, which, for the first time, enhances the model's performance beyond the original. Our approach begins with a theoretical analysis of the relationship between low-rank factorization and model optimization objectives, establishing a precise perturbation range for matrix factorization errors on model performance. This challenge is then reformulated as a numerical rank deficiency problem with inequality constraints and develop a joint objective that simultaneously addresses factorization error and model performance.
Based on the above analysis, we propose two optimization algorithms: \textbf{a lossless optimization algorithm} that maximizes model accuracy while ensuring compression, and \textbf{a compact optimization algorithm} that minimizes model size while preserving performance. These algorithms do not require fine-tuning and can directly compress numerous deep models to achieve lossless results. Our methods demonstrate robust efficacy across various vision and language tasks. For example, the compressed model reduced by 70\% on ResNext50 outperforms the original. Our code will be made public.
\end{abstract}    
\vspace{-0.4cm}
\section{Introduction}
\vspace{-0.1cm}
Deep neural networks have excellent performance in language and vision tasks. A common problem is the significant increase in the number of parameters, which brings challenges to deployment and inference.
Matrix factorization is a common and promising compression method in the community. The idea of matrix factorization is to decompose the weight matrix into two or more smaller matrices and use these two small matrices in actual storage and calculation. Compared with other compression methods such as pruning and quantization, matrix factorization has solid mathematical theoretical support and can fully preserve the intrinsic structure and potential information of the data.
Traditional matrix factorization methods include factorization, Singular Value Decomposition (SVD)~\cite{kalman1996singularly}, and eigenvalue decomposition. When matrix factorization is applied to neural networks, these methods are extended to CANDECOMP/PARAFAC (CP)~\cite{SongAutomated}, Tucker decomposition of kernel tensors. These methods are friendly to linear layers and are often used to compress neural networks.
\begin{figure}
\centering
\includegraphics[width=1\linewidth]{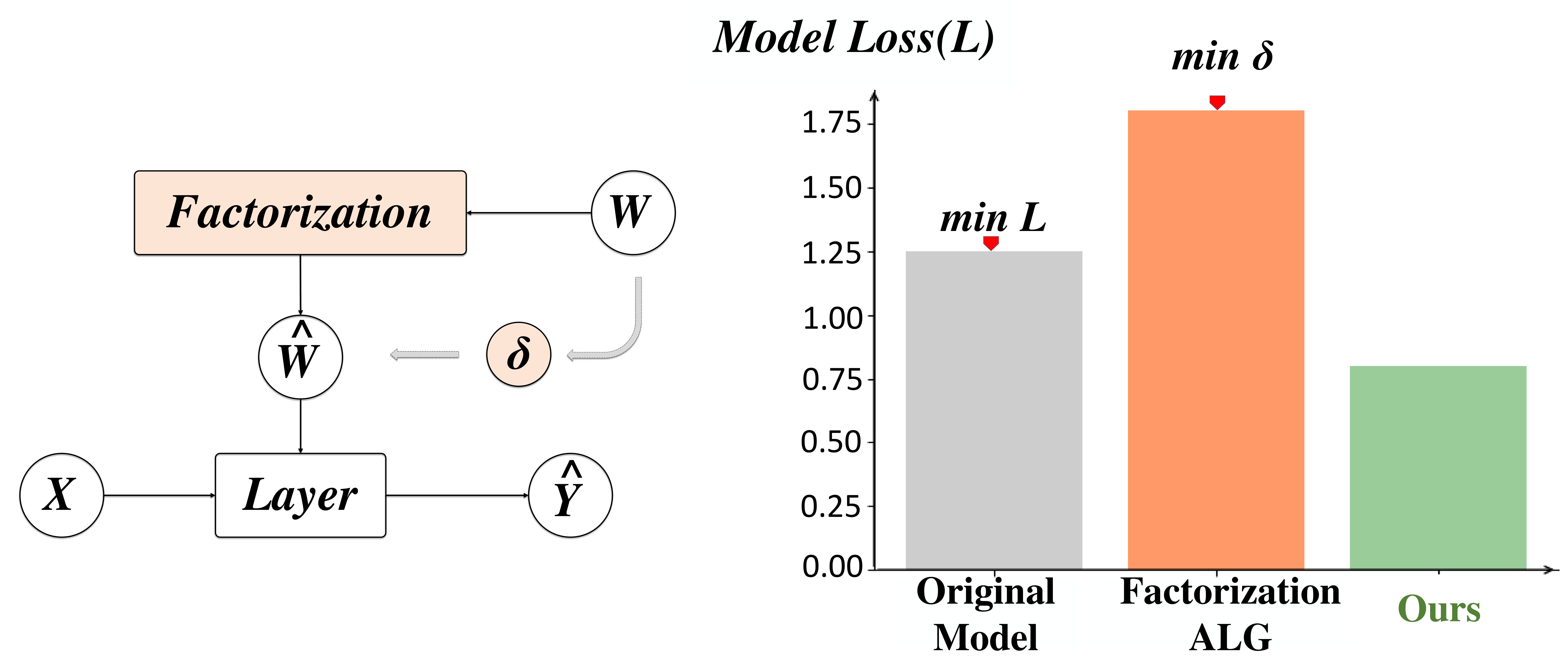}
\vspace{-0.6cm}
\caption{The left subfigure shows the process of factorization, $\delta$ is the noise error introduced by the factorization. The right subfigure shows the Loss comparison between our algorithm and existing factorization algorithms. Our algorithm factorizes models losslessly. $L$ is the model loss.}
\vspace{-0.6cm}
\label{fig1}
\end{figure}

There are two main types of existing post-training low-rank compression schemes. The first scheme is to directly decompose the existing model weights, such as SVD, CANDECOMP/PARAFAC, Tensor-Train (TT), Tucker decomposition, etc., but this type of scheme will lead to increased loss and poor model performance, such as a 5-10 times increase in loss.
To improve this scheme, the second scheme is proposed, that is, after direct weight factorization, different strategies are used to fine-tune the model on the entire dataset. Finetuning is to decompose the model and retrain it using the entire data set. Several studies have contributed to this second scheme. For instance,
Yu et al.~\cite{yu2017compressing} considered weight structure information and combined low-rank weight matrix and feature map reconstruction to reduce fully connected layer parameters.
Xu et al.~\cite{xu2019trained} integrated low-rank approximation and regularization into training with less performance loss.
Yang et al.~\cite{yang2020learning} proposed SVD training, decomposing each layer into a full-rank form, and then directly training the decomposed weights.
Hsu et al.~\cite{hsu2022language} used Fisher information to measure the importance of model parameters and weight them into singular value decomposition.
Yu et al.~\cite{yu2023compressing} adaptively determined the structure of the compression model using a small number of training samples, thereby compressing the output features of each linear layer. 
Kim et al.~\cite{Kim2015CompressionOD} recovered the lost precision through Tucker decomposition on the nuclear tensor and fine-tuning.
Zhang et al.~\cite{zhang2023lightweight} used multiple low-rank matrices to approximate a complete gated recurrent unit weight matrix and retrain.

\begin{figure}
\centering
\includegraphics[width=1\linewidth]{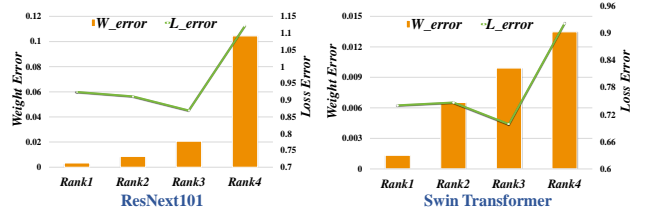}
\vspace{-0.6cm}
\caption{Weight error and loss values of different models, where \textit{Rank1}$>$\textit{Rank2}$>$\textit{Rank3}$>$\textit{Rank4}. When the weight error $\delta$ is the lowest, the model loss $L$ does not reach the minimum value.}
\label{fig_wl}
    \vspace{-0.4cm}
\end{figure}

Although this scheme can improve the model performance when compressing the model, it requires a large amount of fine-tuning data and the fine-tuning process can be time-consuming. It is also not possible to compress the model losslessly, i.e., whether fine-tuned or not, the loss value of the compressed model is always greater than the original model value.
This is because the optimization processes of matrix factorization and model learning are different and separate. These methods follow the \textit{``Factorization + Finetuning''} paradigm. The optimization goal of factorization is to minimize the approximate weight error $\delta$, while the optimization goal of model learning is to minimize the loss function $L$. According to the \textit{``Factorization + Finetuning''} paradigm, the factorization and fine-tuning processes are completed in two stages. Therefore, the approximate error and model loss are separate and have no direct relationship, that is, when the approximate error is minimized, the model loss is not necessarily the lowest, as shown in Figure \ref{fig_wl}.
Unlike previous work, our method considers the joint low-rank factorization and model learning objectives, does not require training and fine-tuning, and establishes a joint optimization process, thereby compressing the model losslessly for the first time, as shown in Figure \ref{fig1}.

Based on the above analysis, we propose a lossless joint low-rank factorization strategy without fine-tuning. Our strategy first establishes the connection between the optimization objectives of low-rank factorization and model learning in theory.
The theoretical limitations in practice are then converted into inequality constraints for the optimization problem.
Specifically, we determine the perturbation range of the loss caused by the factorization error in each layer of the model. Taking this perturbation range as the calculus neighborhood, the connection between the low-rank factorization and the optimization of the original model is mathematically established. Then by imposing error constraints, the factorization optimization problem is converted into a numerical rank-defect optimization problem under inequality constraints, and a new objective related to model performance is proposed. For different requirements, we design two algorithms to solve this problem, lossless optimization and compact matrix optimization algorithms under numerical rank-defect. The lossless optimization algorithm aims to find the lowest loss model for the current layer under model compression. The compact matrix optimization algorithm aims to find the most compact model for the current layer under lossless conditions. It is worth noting that both algorithms do not require fine-tuning and can obtain lossless layers or models.


In summary, the contributions of this paper are as follows:
1) We mathematically establish the connection between factorization and model learning objectives, and propose a lossless joint low-rank factorization strategy that does not require fine-tuning. 2) We convert the traditional factorization optimization problem into a numerical rank-defect optimization problem under inequality constraints and propose two algorithms for different requirements. 3) Extensive experimental results show that our method can ensure compression while obtaining lossless models.

\section{Background}
\textbf{Neural Networks and Optimization.} We present the analysis of neural networks as composite functions. All our conclusions are independent of the structure of the neural network.
First, for an n-layer neural network model, the loss of the model is optimized according to the following formula
\begin{small} 
\begin{equation}
\begin{aligned} & \begin{aligned}
\min_{W}f(W)=\mathbf{E}_{Sample}\ell(W,Sample)=\frac{1}{m}\sum_{(x_{i},y_{i})\in\mathbb{D}}\ell(W,x_{i},y_{i})\end{aligned},\\
&\ell(W,x_{i},y_{i})=L(model_{n}(x_{i},W),y_{i}),\\  & model_{n}=h_1(h_2(h_3(h_4(\cdots(h_{n+1},w_{n})\cdots,w_4),w_3),w_2),w_1),\end{aligned}
\label{eq1}
\end{equation}\end{small}

\noindent where $f(\cdot)$ represents the loss of the model on a dataset, $\mathbf{E}$ stands for expectation, $m$ is the size of the dataset, $\ell(\cdot)$ is the loss function for a sample, and $(x_i,y_i)$ denotes a sample in the dataset along with its corresponding label, $L(\cdot)$ represents the loss function, such as the cross-entropy function; $h_i$, with $i\in[1,...,n]$ represents the ($n-i+1$)th layer in the neural network, $W = (w_n^T,w_{n-1}^T,\cdots,w_1^T)^T$, where $w_i$ is the parameter in $h_i(\cdot)$, and for the reason of a unified format, $h_{n+1}$ denotes the sample $x$.

\textbf{Low-rank Factorization and Optimization.} Low-rank factorization can be adopted to reduce redundancy in weights.
For the weight matrices $W\in\mathbb{R}^{N\times M}$, the low-rank factorization is achieved by two low-rank matrices
\begin{small} 
\begin{equation}
W\approx LR^{T},
\label{eq2}
\end{equation}\end{small}

\noindent where $L\in\mathbb{R}^{N\times k}$, $R\in\mathbb{R}^{M\times k}$, $k$ is the rank of $W$, denoted as an integer between $1$ and $min(N,M)$.
Given input data $x\in\mathbb{R}^{1\times N}$, a linear layer in the neural network can be represented as
\begin{small} 
\begin{equation}
Y=Wx+b\approx \widehat{W}x+b= LR^Tx+b,
\label{eq3}
\end{equation}\end{small}

\noindent where $b$ is the bias, $Y$ is the output of the linear layer, and the factorization of $W$ is obtained from Eq.~\ref{eq1}. With Eq.~\ref{eq2}, we can store and compute $L$ and $R$ instead of $W$. The total number of parameters of $L$ and $R$ is $Nk+Mk$. The reduced parameters and computation are $NM-(Nk+Mk)$. When the weight matrix $M=N$, the rank $k$ is less than $0.5M$ and the model size will be reduced. Similarly, for singular value decomposition, $W$ is approximated as $ \widehat{W}=USV^T$, where $U$ and $V$ are orthonormal, and $S$ is diagonal.

The optimization objective of the low-rank factorization is to minimize the Frobenius norm under the rank is most $k$
\begin{small} 
\begin{equation}
\min_{L,R}||W-\widehat {W}||_{F},
\label{eq4}
\end{equation}\end{small}

\noindent it implies that we can find a basis set for an optimal $k$ rank approximation in a greedy way. 

As shown in Eq.~\ref{eq1} and Eq.~\ref{eq4}, the optimization objective of low-rank factorization is different from the model optimization objective. Although this low-rank optimization objective can best approximate the weight matrix, it does not ensure that the model achieves the lowest loss value. The \textit{``Factorization + Finetuning''} paradigm separates the two optimization processes, resulting in reduced model performance.
\section{Lossless Joint Factorization Strategy}
\textbf{Theoretical Optimization.} The key to lossless factorization optimization is how to analyze low-rank factorization in model optimization. As shown in Figure~\ref{fig1}, mathematically, the nature of low-rank factorization is a process that introduces noise to the weight parameters in the original model or layer. After factorization, for a sample, the model loss $\bar{\ell}$ during inference is reformulated as the following equation

\begin{small}
\vspace{-0.4cm}
\begin{align}
\bar{\ell}_k(w,x_{j},y_{j})  =L(h_1(h_2(\cdots h_{n}(x_i, w_{n}+\delta^k_{n})\cdots,\nonumber  \\w_2+\delta^k_2),w_1+\delta^k_1), y_{i}),
\end{align}
\end{small}

\noindent  where $\delta^k_{i},i\in\left\lbrack1,\cdots,n\right\rbrack$ denotes the noise error on the weights after the rank $k$ factorization. The error $\delta$ caused by low-rank factorization ultimately leads to changes in model loss. This error can be seen as an increment in weight $w$.

\vspace{-0.1cm}
\begin{small} 
\begin{lemma}\label{lemma1}
The weight increment of the loss function at a certain point can be estimated by the sum of the products of each partial derivative and a small change in the weight variable.
\end{lemma}\end{small}

\vspace{-0.1cm}
\noindent Lemma~\ref{lemma1} is the expression of the concept of total differential in weight factorization in neural networks.
Then, through the definition of total differential~\cite{matthews2012vector} and Lemma~\ref{lemma1}, for the differentiable function ${\ell}$ with the small variable $\delta$, the following equation is established
\begin{small} 
\begin{equation}
\bar{\ell}_k(w,x_{i},y_{i})-\ell(w,x_{i},y_{i})=\sum_{i=1}^{n}\frac{\partial\ell_k}{\partial w_i}\cdot\delta^k_{i},
\label{eq6}
\end{equation}\end{small}

\noindent where $\cdot$ is the inner product. For the loss on the entire dataset, the optimization objective is written as
\begin{small} 
\begin{equation}\begin{aligned}\operatorname*{min}_{\delta\in \Delta}\bar{f}(w)-f(w) & =\frac{1}{m}\sum_{i=1}^{n}\sum_{(x_{j},y_{j})\in\mathbb{D}}\frac{\partial\ell_k}{\partial w_i}\cdot\delta^k_{i},
\label{eq7}
\end{aligned}\end{equation}\end{small}

\noindent where $\bar{f}(w) = \frac{1}{m}\sum \bar{\ell_k}(\cdot)$, $\Delta$ is the full set of $\delta$. This optimization objective incorporates the noise error from the factorization into the model loss. Eq.~\ref{eq7} needs to satisfy the \textit{loss continuity and differentiability}, and the \textit{continuity of the partial derivatives}. The most important thing is that the change is small enough because when the change is small enough, the total differential provides a good approximation. This will be discussed in the next section.

Eq.~\ref{eq7} expresses the loss difference before and after the factorization as the product of the gradient vector and the noise vector. When the inner product is negative when the two vectors are in opposite directions, i.e. $\sum_{(x_{j},y_{j})\in\mathbb{D}}\frac{\partial\ell}{\partial w_i}\cdot\delta^k_{i}<0$, the model loss after factorization will be less than the original model. Thus, the goal of lossless joint low-rank factorization is achieved. 

\textbf{Theoretical Conditions and Practical Mapping.} For the theoretical validity of Eq.~\ref{eq7}, specific conditions must be met, particularly regarding the requirements for mathematical differentiation. Since the loss in neural networks is continuously differentiable with a continuous gradient, we focus on the constraints associated with changes in $w$. It is necessary to analyze the size of the neighborhood when applying differential expansions to differentiable functions.
\vspace{-0.3cm}
\begin{small}
\begin{lemma}
The local linear approximation of the change in function value is only valid if the change in weights is small enough.
\label{lenma2}
\end{lemma}\end{small}
\vspace{-0.3cm}
Corresponding to Lemma~\ref{lenma2}, usually in mathematics, a neighborhood is a sufficiently small range $\epsilon$ around the point of expansion. So in theory, Eq.~\ref{eq7} needs to satisfy the neighborhood
$U_{\delta^k}\left(x_i\right): |\delta^k|<\epsilon$.

In practice, how to obtain this neighborhood is the prerequisite for the effectiveness of our method. Therefore we calculated the gap between the loss values in Eq.~\ref{eq7} under theoretical analysis and in practical engineering.
\begin{small} 
\begin{equation}
U_{\delta^k}\left(x_i\right): |{\ell}(w \pm \delta^k_{i} ,x_{i},y_{i}) -(\ell(w,x_{i},y_{i})+\sum_{i=1}^{n}\frac{\partial\ell_k}{\partial w_i}\cdot\delta^k_{i})|,
\label{eq8}
\end{equation}\end{small}

\noindent The meaning of the above equation is to verify under what range of neighborhoods the theoretical derivation of Eq.~\ref{eq6} can be implemented in practical experiments. The left side of the minus sign is the loss under practical engineering for a specific layer with parameters added to the factorization noise $\delta^k$, and the right side is the loss in the theoretical analysis for the weight gradient perturbation, corresponding to Eq.~\ref{eq6}. We consider using the above equation to determine what the rank is, the noise $\delta^k$ caused by the factorization can make this difference small enough so that Eq.~\ref{eq6} or Eq.~\ref{eq7} holds. Through multiple rounds of experiments, we find that this theoretical and practical difference is sufficiently small to be less than 0.0001 when the noise introduced by the decomposition is $\delta^k \leq \epsilon \in \mathcal{O}(10^{-3})$. This ensures that our analysis is valid. Please see the experimental section for specific values and schemes. The determination of the neighborhood ensures the representation of low-rank factorization in total differentials, and we map the theory of total differentials into practice.

\textbf{Practical Constraints.} By mapping the total differential theory to practice, lossless joint low-rank factorization needs to ensure two conditions called the compression condition and the lossless condition. The compression condition is generated to efficiently compress the model, i.e., the number of parameters of the original matrix in Eq.~\ref{eq2} is smaller than the number of parameters of the approximation matrix, $0<k<NM/(N+M)$. The lossless condition is generated to efficiently reduce the loss of the model, which corresponds to the condition for the differentiation to hold. The noise $\delta^k$ after factorization should be less than $\epsilon$, i.e. $\{\Vert w_{ij}-l_{ij}r_{ij} \Vert \}_{i,j} \leq\epsilon$, where  represented by the set of matrix elements, $l_{ij}r_{ij} $ represents the $i,j$th element of the approximation matrix $LR^T$. Due to the directionality of the gradient vector, we use the matrix $F$ norm for calculation. This restriction means that every element in the difference set of all original matrices and approximate matrices needs to be smaller than $\epsilon$. 
Lossless low-rank factorization optimization needs to satisfy these two inequality constraints at the same time, which is updated to the following formula

\begin{subequations}\label{eq:ctr_shale1}
\vspace{-0.4cm}
\begin{align}
\operatorname*{min}_{\delta^k\in \Delta}\bar{f}(w)-f(w) & =\frac{1}{m}\sum_{i=1}^{n}\sum_{(x_{j},y_{j})\in\mathbb{D}}\frac{\partial\ell}{\partial w_i}\cdot\delta^k_{i},  \ \ \tag{\ref{eq:ctr_shale1}}
\end{align} 
\vspace{-0.3cm}
\begin{alignat}{2}
\text{s.t.  } & U_{\delta^k}: \{\Vert w_{ij}-l_{ij}r_{ij} \Vert_F \}_{i,j} \leq\epsilon , \ \forall i,j ,  \\
    &0<k<\frac{NM}{N+M}.
\end{alignat}
\end{subequations}

It is worth noting that our goal is still to compress the model. So the compression condition should be satisfied before the lossless condition.
Secondly, our algorithm is flexible, and the lossless condition expression can also be replaced by other decompositions, such as $\widehat{w}=usv^T$. However, when replaced by other decompositions, the number of parameters and time may increase. Eq.~\ref{eq:ctr_shale1} is the simplest expression of the effective factorization method.

Theoretical optimization ensures the possibility of lossless decomposition, while practical mapping provides a viable solution for lossless compression. The lossless strategy does not rely on assumptions; the full differential analysis is theoretically guaranteed and validated through practice, and the noise constraints are computed based on specific models rather than assumptions.

\textbf{Gradient and Higher-order Terms.}
When optimizing Eq.~\ref{eq:ctr_shale1}, we also need to analyze gradients and higher-order terms. Theoretically, for a well-trained model, the expectation of $\ell(\cdot)$'s gradient for parameters is ideally zero, i.e., for the $\sum_{(x_j,y_j)\in \mathbb{D}} \frac{\partial \ell}{\partial w}$ components, $\frac{\partial \ell}{\partial w_i} = 0$. But in practice, we found that almost no model weight gradient is 0. This also ensures the feasibility of our algorithm. Moreover, the second-order term is influenced by the infinitesimal higher-order term. So we focus on first-order elements, and higher-order terms will not cause major performance changes when they are ignored.

\vspace{-0.1cm}

\begin{algorithm} 
    \small
	\caption{: Lossless Optimization Algorithm}     
	\label{alth1}       
	\begin{algorithmic}[1]

		\Require Neural network $M$, lossless rank list $A$.   
		\Ensure  Loss-minimization model $\hat{M}$  after factorization.   

		\For {$Layer_a$ in $M$}
			\State Compute the maximum rank $rank_{max}$ in compression conditions.
			\If{$Layer_a$ is decomposed} \State continue
			\EndIf

			\For {$c = 1, 2, 3, ..., rank_{max}$}
				\State Initialize $L_c$, $R_c$ {\Comment{Temporary matrix}}
				\State Compute the error $\delta_i$ of $h_{i+1}$ under $rank_{c}$ level on dataset
				\If{$\{\Vert w_{ij}-l_{ij}r_{ij} \Vert_F \}_{i,j} \leq\epsilon , \ \forall i,j$ }
                    \If{$\frac{\partial\ell}{\partial w_i} \cdot \delta_i < 0$}
                        \State \{$L_c$, $R_c$, $Loss$\} append to list $A$
				    \EndIf
				\EndIf
                \State Optimize and Update $L_c$, $R_c$
			\EndFor
            \State Search for the minimized Loss in the List $A$.
            \State $W_a$ = $L_a$ $R_a$  {\Comment{Final decomposition matrix}}
            \State Return $\hat{Layer_a}$ 
		\EndFor
  
       \State Return $\hat{M}$  
        
	\end{algorithmic} 

\end{algorithm} 

 \vspace{-0.5cm}
\section{Lossless and Compact Optimization}
Based on the above analysis and optimization strategies, we propose two algorithms with different strategies, lossless and compact matrix optimization algorithms. Lossless and compressive factorization algorithms are aggressive probabilistic algorithms that do their best. The lossless factorization algorithm focuses on minimizing the loss function value while keeping the model size no larger than a full-precision model. The compressed matrix algorithm aims to minimize the model size while ensuring that the loss function value is not larger than that of the full precision model. Both algorithms are greedy algorithms to achieve the goal of minimizing the loss function value or model size. 

In Algorithm~\ref{alth1} and Algorithm~\ref{alth2}, our goal is to find the noise introduced by low rank that is opposite to the gradient direction, so that the inner product of the two is negative and reduces the model loss.
In Algorithm~\ref{alth1}, the priority is to achieve a low loss function value to obtain the best performance factorization model. 
In Algorithm~\ref{alth2}, priority is given to choosing a lower rank to obtain a smaller factorization model. 

\vspace{-0.1cm}

\begin{algorithm} 
    \small
	\caption{: Compact Optimization Algorithm}     
	\label{alth2}       
	\begin{algorithmic}[1] 
		\Require Neural network $M$.   
		\Ensure  Rank-minimization model $\hat{M}$ after factorization.    

		\For {$Layer_a$ in $M$}
			\State Compute the maximum rank $rank_{max}$ in compression conditions
			\If{$Layer_a$ is decomposed}
				\State continue
			\EndIf

			\For {$c = 1, 2, 3, ..., rank_{max}$}  \Comment{Binary search optimization}
				\State Initialize $L_c$, $R_c$ {\Comment{Temporary matrix}}
				\State Compute the error $\delta_i$ of $h_{i+1}$ under $rank_{c}$ level on dataset
				\If{$\{\Vert w_{ij}-l_{ij}r_{ij} \Vert_F \}_{i,j} \leq\epsilon , \ \forall i,j$ }
                    \If{$\frac{\partial\ell}{\partial w_i} \cdot \delta_i < 0$}
                        \State  $W_a$ = $L_a$ $R_a$ {\Comment{Final matrix}}
				    \EndIf
				\EndIf
                \State Update $L_c$, $R_c$
			\EndFor
            \State Return $\hat{Layer_a}$ 
		\EndFor
  
       \State Return $\hat{M}$  
        
	\end{algorithmic} 
\end{algorithm}

   \vspace{-0.2cm}
\textbf{Generalization error upper bound.}
Although the algorithm will reduce loss, we need to discuss the upper bound of the generalization error under the algorithm. Lemma~\ref{lemma3} guarantees the existence of our algorithm so that the weights can be decomposed. We can prove it by Pointwise Hypothesis Stability~\cite{Fu2022OnTE}.

\vspace{-0.2cm}
\begin{small}
\begin{lemma}
The upper bound of the generalization error R decreases with decreasing effective rank. After reducing the effective rank to a certain value, the upper bound of the generalization error will increase as the rank decreases further.
\label{lemma3}
\end{lemma}\end{small}

\vspace{-0.2cm}
\noindent \textit{\textbf{proof:}} Consider $A$ as our algorithm and $(X,Y)$ as a dataset of length $s$. In addition, consider the ratio of the effective rank to the original rank as $p$ (where $1$ -- $p$ is the sparsity parameter). The generalization error upper bound can be calculated by assuming the pointwise stability equation. We have for a constant $T$ with a probability $1$ -- $\beta$,
\begin{small}
\begin{equation}\mathcal{R}(\mathcal{A},X,Y)<\hat{\mathcal{R}}(\mathcal{A})+\sqrt{\frac{T^2+\frac{24T\rho^2}{\lambda_{min}+2(1-p)}}{2s\beta}},
\end{equation}
\end{small}

\noindent where $\hat{\mathcal{R}}(\mathcal{A})$
represents the emperical error, and $\lambda_{min}$ represents the minimum eign-value of the loss Hermitian matrix. For the loss Hermitian matrix~\cite{Sagun2016EigenvaluesOT}, since the model has been trained, $\lambda_{min} \approx 0$. Based on the above equation, we can observe that when the effective rank ratio $p$ is low and the sparsity (1 -- $p$) is relatively high, the generalization error decreases with increasing sparsity. However, as the effective rank decreases and the sparsity increases, there comes a point where the number of trainable parameters is much lower than the number required to represent the distribution of the dataset, resulting in underfitting. Therefore, there is an optimal effective rank that allows the weights to be decomposed. This ensures that the algorithm is effective.

\section{Experiments}
\subsection{Datasets and Details} 
\textbf{Datasets.} The ImageNet-1K dataset~\cite{krizhevsky2017imagenet} consists of 1.28M training and 50K validation images. ImageNet-1K is usually used as the benchmark for model compression.
SWAG dataset~\cite{zellers2018swag} consists of 113K multiple-choice questions about grounded situations. The Stanford Question Answering Dataset (SQuAD)~\cite{rajpurkar2016squad} is a collection of question-answer pairs derived from Wikipedia articles. In SQuAD, the correct answers to questions can be any sequence of tokens in the given text. MNLI~\cite{williams2017broad} is a dataset for natural language reasoning tasks. Its corpus is a collection of textual implication annotations of sentences through crowdsourcing. The task is to predict whether the premise sentence and the hypothesis sentence are logically compatible (entailment, contradiction, neutral). CoNLL-2003~\cite{sang2003introduction} is a named entity recognition dataset released as a part of CoNLL-2003 shared task: language-independent named entity recognition. The data consists of eight files covering two languages: English and German. 

\textbf{Details.} Our method is implemented in Pytorch,  and does not require fine-tuning and retraining. Following existing compression work, we use the ImageNet validation set as the calibration set, computing the gradient once to check the gradient values rather than updating the weights, and based on this, we determine the decomposition rank for each layer. Use the same optimization settings for all experiments in this paper and avoid any hyperparameter filtering to ensure a fair comparison. The experiments are run on a single NVIDIA V100 GPU.

\begin{table*}[!ht]
    \centering
        \renewcommand{\arraystretch}{1}
    \setlength{\tabcolsep}{8.3pt}
    \caption{Accuracy and loss results on Imagenet for popular Shallow or Deep models. -Ours(L) is the lossless factorization result, -Ours(C) is the compression factorization algorithm result. Bold is the best except for the original model accuracy.}
    \vspace{-0.2cm}
    \scalebox{0.74}{
    \begin{tabular}{c|c|c|c|c|c|c|c|c|c}
    \hline
        \textit{\textbf{Shallow Models}} & \textbf{Top-1} & \textbf{Top-5} & \textbf{Loss} & \textbf{Drop Rate} & \textit{\textbf{Deep Models}} & \textbf{Top-1} & \textbf{Top-5} & \textbf{Loss} & \textbf{Drop Rate} \\ \hline
        VGG-19\_BN~\cite{karen2014very} & 74.218  & 91.842  & 1.042591  & ~ & ResNext101\_32x4d~\cite{xie2017aggregated} & 79.312  & 94.526  & 0.926616  & ~ \\ \hline
        VGG-19\_BN-SVD & 74.222  & 91.864  & 1.042603  &  \textcolor{red}{-2.00\%} & ResNext101\_32x4d-SVD & 78.154  & 94.012  & 0.870145  &  \textcolor{red}{-75.00\%} \\ \hline
        \rowcolor{gray!20} VGG-19\_BN-Ours(L) & \textbf{74.222}  & \textbf{91.892}  & \textbf{1.021449}  &  \textcolor{red}{-2.00\%} & ResNext101\_32x4d-Ours(L) & \textbf{78.160}  & \textbf{94.018} & \textbf{0.869111}  &  \textcolor{red}{-75.00\% }\\ \hline
        \rowcolor{yellow!20}VGG-19\_BN-Ours(C) & 74.112  & 91.715  & 1.042593  & \textcolor{red}{\textbf{ -43.00\%}} & ResNext101\_32x4d-Ours(C) & 76.308  & 91.482  & 0.924511  &  \textcolor{red}{\textbf{-81.00\%}} \\ \hline
        Inception V3~\cite{szegedy2016rethinking} & 69.538  & 88.654  & 1.829029  & ~ & DenseNet169~\cite{huang2017densely} & 75.600  & 92.806  & 0.997792  & ~ \\ \hline
        Inception V3-SVD & 66.262  & 87.258  & 1.554306  &  \textcolor{red}{-68.00\%} & DenseNet169-SVD & 75.426  & \textbf{92.802}  & 1.001437  &  \textcolor{red}{-2.00\%} \\ \hline
        \rowcolor{gray!20} Inception V3-Ours(L) & \textbf{67.872}  & \textbf{87.866}  & \textbf{1.526568 } &  \textcolor{red}{-68.00\%} & DenseNet169-Ours(L) & \textbf{75.446}  & 92.790  & \textbf{0.971887}  &  \textcolor{red}{-2.00\%} \\ \hline
        \rowcolor{yellow!20}Inception V3-Ours(C) & 60.784  & 84.242  & 1.823540  &  \textcolor{red}{\textbf{-82.00\%}} & DenseNet169-Ours(C) & 74.790  & 92.572  & 0.994877  &  \textcolor{red}{\textbf{-52.00\%}} \\ \hline
        ResNext50\_32x4d & 77.618  & 93.698  & 0.940085  & ~ & DenseNet201 & 76.896  & 93.370  & 0.926975  & ~ \\ \hline
        ResNext50\_32x4d-SVD & 76.300  & 93.112  & 0.934977  &  \textcolor{red}{-68.00\% }& DenseNet201-SVD & \textbf{76.846}  & 93.300  & 0.928765  &  \textcolor{red}{-2.00\%} \\ \hline
        \rowcolor{gray!20}ResNext50\_32x4d-Ours(L) & \textbf{77.008}  & \textbf{93.420}  & \textbf{0.920947}  &  \textcolor{red}{-68.00\%} & DenseNet201-Ours(L) & 76.768  & \textbf{93.304}  & \textbf{0.919823}  &  \textcolor{red}{-2.00\%} \\ \hline
        \rowcolor{yellow!20}ResNext50\_32x4d-Ours(C) & 76.284  & 93.114  & 0.938988  &  \textcolor{red}{\textbf{-78.00\%}} & DenseNet201-Ours(C) & 76.582  & 93.210  & 0.924847  &  \textcolor{red}{\textbf{-25.00\%}} \\ \hline
    \end{tabular}} \label{tab1}
        \vspace{-0.4cm}
\end{table*}

\begin{table}[ht]  
\renewcommand{\arraystretch}{1}  
\setlength{\tabcolsep}{3.7pt}  
\caption{  
Comparison of the algorithm with existing decomposition methods on Imagenet. Data size indicates the amount of data required by the algorithm. The bold ones are the best except for the original model.}  
\vspace{-0.1cm}  
\scalebox{0.76}{  
\begin{tabular}{c|c|c|c|c|c}  
\hline  
\rowcolor[HTML]{FFFFFF}   
\textbf{VGG-16}     & \textbf{Top-1} & \textbf{Top-5} & \textbf{Loss} & \textbf{Cost Time} & \textbf{Data Size} \\   
\hline  
\rowcolor[HTML]{FFFFFF}   
Original Model      & 71.586        & 90.374        & 1.144342      & -                 & -                  \\   
\hline  
\rowcolor[HTML]{FFFFFF}   
SVD\_ft             & 71.546        & 90.364        & 1.145468      & 314.985 min       & 140 G              \\   
\hline  
\rowcolor[HTML]{FFFFFF}   
Tai et al.~\cite{tai2015convolutional} & - & 90.310 & - & - & 140 G \\   
\hline  
\rowcolor[HTML]{FFFFFF}   
Kim et al.~\cite{Kim2015CompressionOD} & - & 89.400 & - & - & 140 G \\   
\hline  
\rowcolor[HTML]{FFFFFF}   
Zhang et al.~\cite{zhang2023lightweight} & 71.550 & 90.350 & 1.233013 & 433.115 min & 140 G \\   
\hline  
\rowcolor{gray!20}  
Ours(L) & 71.484 & \textbf{90.378} & \textbf{1.139378} & 8.287 min & \textbf{6.4 G} \\   
\hline  
\rowcolor{yellow!20}  
Ours(C) & \textbf{71.560} & 90.364 & 1.144460 & \textbf{7.762} min & \textbf{6.4 G} \\   
\hline  
\end{tabular}}  
\label{tabCV}  
\vspace{-0.5cm}  
\end{table}  

\begin{table}[!hb]
    \centering
    \renewcommand{\arraystretch}{1.2}
    \setlength{\tabcolsep}{8.8pt}
    \caption{Comparison with other compression methods. Our method enables lossless factorization with a higher drop rate.}
\vspace{-0.2cm}
\scalebox{0.8}{
    \begin{tabular}{c|c|c|c|c}
    \hline
        \textbf{Model} & \textbf{Top-1} & \textbf{Top-5} & \textbf{Loss} & \textbf{Drop rate }\\ \hline
        ResNext50\_32x4d & \textbf{77.618}  & \textbf{93.698}  & 0.940085  & ~ \\ \hline
          + ACIQ \cite{banner2018aciq} & 77.140 & 93.382 & 0.940222  & \textcolor{red}{ $-$  73.00\%} \\ \hline
         \rowcolor{gray!20} + Ours(L) & 77.008  & 93.420  & \textbf{0.920947}  & \textcolor{red}{ $-$  68.00\%} \\ \hline
         \rowcolor{yellow!20} + Ours(C) & 76.284  & 93.114  & 0.938988  & \textcolor{red}{ $-$  78.00\% }\\ \hline
    \end{tabular}} \label{tabother}
    \vspace{-0.1cm}
\end{table}

\subsection{Lossless and Comparative Experiments}
\subsubsection{Lossless Experiments}
In the lossless experiments, we primarily validate the lossless decomposition effects of two algorithms using models of different scales. As a comparative baseline, we select the SVD method, as SVD focuses on optimizing the gaps between factorized matrices while neglecting model optimization. Our optimization goal takes into account both factorization matrix optimization and model optimization. Table \ref{tab1} shows the accuracy and loss results of our algorithm on Imagenet. We factorize all linear weight matrices in the visual classifier. We conduct experiments on models of different types and depths. For models of different depths, our lossless factorization and compression factorization methods can perform lossless factorization on both shallow and deep models.  Compared with deep models, shallow models have less parameter redundancy and are more difficult to compress. Our algorithm is also effective for shallow models. 
Although the SVD algorithm can approximate the original matrix and reduce the approximation error, it will increase the loss of the model. This also shows that in \textit{``Factorization + Finetuning''}, reducing the approximation error and reducing the loss value are not linearly related. Our algorithm establishes the relationship between factorization and model learning and establishes a lossless joint optimization goal. The loss value of our algorithm is not only smaller than the SVD method, but also smaller than the original model loss, and the accuracy is comparable to or even higher than the original model. The lossless compression results of different models also prove the generalization of our algorithm. 
Secondly, the change in loss is related to the weight gradient and the noise amplitude. When the amplitude is larger, the loss decreases more, such as ResNext101\_32x4d. 
In ResNext101\_32x4d, the loss of the model after lossless factorization is reduced compared with the original model, and the compression is completed. Compared with the original model, the compact optimization algorithm (Ours-C) can reduce 81\% of parameters and is less than the loss of the original model. Since the algorithm is mathematically interpretable, it is independent of the model architecture. 

\subsubsection{Comparison Experiments}
In comparative experiments, we conduct experiments on tasks in different fields to fully verify the effectiveness.

\begin{figure}
\centering
\includegraphics[width=8.5cm,height=2.7cm]{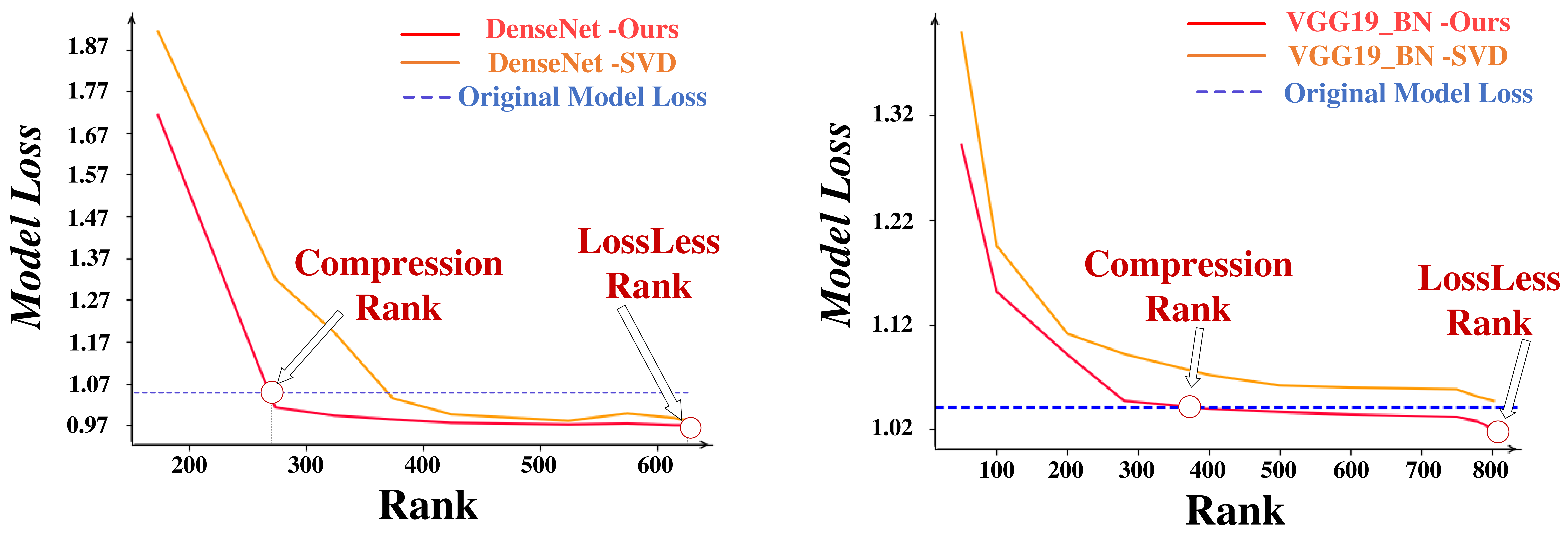}
\vspace{-0.6cm}
\caption{Loss performance of our algorithm on deep DenseNet169 and shallow VGG-19 model. Our algorithms have lower losses than the original model under the condition that $0<Rank<\frac{NM}{N+M}$ is satisfied.}
\vspace{-0.5cm}
\label{fig_rank}
\end{figure}

\textbf{Image Classification.}
Table \ref{tabCV} shows the comparison of our algorithm with existing decomposition methods on Imagenet. We also fine-tune some methods (SVD\_ft). The experiments show that our lossless algorithm achieves the lowest loss, and the compression algorithm achieves the highest accuracy while ensuring that the loss remains almost unchanged. Unlike these methods, our algorithm only requires the gradient direction and does not need to use the gradient for update calculation. Our algorithm can compress the model losslessly while running in the shortest time and using the least data.
Figure \ref{fig_rank} shows our lossless factorization strategy in inference when $0<rank<\frac{NM}{N+M}$ is satisfied. The red line in the figure represents the loss-rank curve of our algorithm, and the orange line represents the loss-rank curve of the SVD algorithm. As the rank decreases, the loss shows an increasing trend. The intersection point of the model loss straight line and our algorithm curve is the lowest rank when the model loss is almost the same. The compression factorization optimization algorithm does its best to compress the model while ensuring that the loss remains unchanged. Under the constraints of compression conditions, the lowest point of our algorithm is the minimum loss of the curve. The lossless factorization method does its best to reduce model loss while ensuring compression. The different changes in loss curves also correspond to Lemma~\ref{lemma3}. The algorithm finds an effective rank less than full rank so that the weights can be decomposed.

Table \ref{tabother} shows the comparison between our algorithm and quantization method. ACIQ uses int8 for quantization, which improves the loss of the model. Compared with ACIQ, our algorithm has a higher compression rate at a similar running time. Our algorithm can choose different factorization strategies as needed.

\begin{table*}[!hb]
\renewcommand{\arraystretch}{1}

\setlength{\tabcolsep}{2.4pt}
\vspace{-0.2cm}
\caption{
Performance of algorithms on language processing tasks. The test set includes SQuAD, MNLI, and CoNLL-2003.}
\vspace{-0.2cm}
\scalebox{0.83}{
\begin{tabular}{ccccccccccccc}
\hline
\multicolumn{1}{c|}{}                                  & \multicolumn{4}{c|}{\textbf{SQuAD}}                                                                                                                                                                      & \multicolumn{4}{c|}{{ \textbf{MNLI}}}                                                                                                                                                                               & \multicolumn{4}{c}{\textbf{CoNLL-2003 NER}}                                                                                                                                                                                        \\ \cline{2-13} 
\multicolumn{1}{c|}{\multirow{-2}{*}{\textbf{Method}}} & \multicolumn{1}{c|}{\textbf{Acc on Val}} & \multicolumn{1}{c|}{\textbf{EM}}      & \multicolumn{1}{c|}{\textbf{F1}}                        & \multicolumn{1}{c|}{\textbf{Loss}}                          & \multicolumn{1}{c|}{\textbf{Acc on Val}}             & \multicolumn{1}{c|}{{ \textbf{Loss on Val}}} & \multicolumn{1}{c|}{\textbf{Acc on Test}} & \multicolumn{1}{c|}{{ \textbf{Loss on Test}}} & \multicolumn{1}{c|}{{ \textbf{Prec.}}} & \multicolumn{1}{c|}{\textbf{Recall}}  & \multicolumn{1}{c|}{{ \textbf{F1}}} & \multicolumn{1}{c}{\textbf{Loss}}                           \\ \hline
\multicolumn{1}{c|}{BERT\_base}                        & \multicolumn{1}{c|}{85.74}              & \multicolumn{1}{c|}{80.4920}           & \multicolumn{1}{c|}{88.1464}                            & \multicolumn{1}{c|}{\textbf{0.4461}}                                 & \multicolumn{1}{c|}{{ 82.77}}  & \multicolumn{1}{c|}{{ 0.0289}}             & \multicolumn{1}{c|}{83.91}              & \multicolumn{1}{c|}{0.0285}                                      & \multicolumn{1}{c|}{89.94}                                    & \multicolumn{1}{c|}{91.69}          & \multicolumn{1}{c|}{{ 90.79}}          & \multicolumn{1}{c}{8.5685}                                 \\ \hline
\multicolumn{1}{c|}{BERT\_base-SVD}                    & \multicolumn{1}{c|}{83.78}            & \multicolumn{1}{c|}{79.0445}         & \multicolumn{1}{c|}{{ 86.8670}} & \multicolumn{1}{c|}{{ 0.5168}}        & \multicolumn{1}{c|}{{ 81.69}} & \multicolumn{1}{c|}{0.0302}                                      & \multicolumn{1}{c|}{82.65}              & \multicolumn{1}{c|}{{ 0.0299}}               & \multicolumn{1}{c|}{{ 89.32}}            & \multicolumn{1}{c|}{90.97}           & \multicolumn{1}{c|}{{ 90.13}}          & \multicolumn{1}{c}{{ 8.6458}}          \\ \hline
\multicolumn{1}{c|}{CANDECOMP \cite{SongAutomated}}                     & \multicolumn{1}{c|}{84.51}               & \multicolumn{1}{c|}{79.9460}           & \multicolumn{1}{c|}{87.5400}                          & \multicolumn{1}{c|}{{ 0.6561}}           & \multicolumn{1}{c|}{81.46}                         & \multicolumn{1}{c|}{0.0340}                                       & \multicolumn{1}{c|}{81.54}               & \multicolumn{1}{c|}{0.0310}                                        & \multicolumn{1}{c|}{{89.32}}            & \multicolumn{1}{c|}{90.15}          & \multicolumn{1}{c|}{89.66}                                 & \multicolumn{1}{c}{9.4560}                                   \\ \hline
\multicolumn{1}{c|}{Zhang et.al.~\cite{zhang2023lightweight}}                        & \multicolumn{1}{c|}{85.02}               & \multicolumn{1}{c|}{80.0120}           & \multicolumn{1}{c|}{88.0445}                          & \multicolumn{1}{c|}{{ 0.5978}}          & \multicolumn{1}{c|}{80.45}                         & \multicolumn{1}{c|}{0.0570}                                       & \multicolumn{1}{c|}{80.89}              & \multicolumn{1}{c|}{0.0742}                                        & \multicolumn{1}{c|}{{ \textbf{90.01}}}   & \multicolumn{1}{c|}{90.46}          & \multicolumn{1}{c|}{{ 89.99}}          & \multicolumn{1}{c}{9.2440}                                   \\ \hline
\rowcolor{yellow!20}\multicolumn{1}{c|}{BERT\_base-Ours(C)}                & \multicolumn{1}{c|}{\textbf{85.67}}     & \multicolumn{1}{c|}{\textbf{80.4257}} & \multicolumn{1}{c|}{\textbf{88.1622}}                   & \multicolumn{1}{c|}{{ 0.4482}} & \multicolumn{1}{c|}{\textbf{82.78}}                & \multicolumn{1}{c|}{{ \textbf{0.0289}}}    & \multicolumn{1}{c|}{\textbf{83.92}}     & \multicolumn{1}{c|}{\textbf{0.0285}}                             & \multicolumn{1}{c|}{{89.30}}            & \multicolumn{1}{c|}{\textbf{90.99}} & \multicolumn{1}{c|}{\textbf{90.13}}                        & \multicolumn{1}{c}{{ \textbf{8.5487}}} \\ \hline

\end{tabular}} \label{tabNLP}
\end{table*}

\begin{table}[ht]
\renewcommand{\arraystretch}{1}
\setlength{\tabcolsep}{6.5pt}
\vspace{-0.2cm}
\caption{Comparison of this algorithm with existing decomposition methods on SWAG under the same drop ratio.}
\vspace{-0.3cm}
\renewcommand{\arraystretch}{1.2}
\scalebox{0.75}{
\begin{tabular}{
>{\columncolor[HTML]{FFFFFF}}c |
>{\columncolor[HTML]{FFFFFF}}c |
>{\columncolor[HTML]{FFFFFF}}c |
>{\columncolor[HTML]{FFFFFF}}c |
>{\columncolor[HTML]{FFFFFF}}c }
\hline
\textbf{BERT\_base  }               &\textbf{ Acc. }                            & {\color[HTML]{333333} \textbf{Loss}}     & \textbf{Cost Time} & {\color[HTML]{333333} \textbf{Data Size}} \\ \hline
Original Model           & \textbf{79.111 }                       & 0.057902                        & -          &  -                                  \\ \hline
SVD\_ft   & 78.442                       & 0.059100                          & 134.649 min       & 27 M                                  \\ \hline
CANDECOMP~\cite{SongAutomated}   & 78.556                       & 0.064031                        & 151.006 min       & 27 M                                   \\ \hline
Zhang et.al.~\cite{zhang2023lightweight}& {\color[HTML]{333333} 79.002} & 0.072234                        & 233.146 min       & 27  M                                      \\ \hline
\rowcolor{yellow!20}
Ours(L)   & 78.566                        & \textbf{0.056642  }                      & 11.413  min       &\textbf{7.6 } M                                      \\ \hline
\rowcolor{yellow!20} 
Ours(C) $\downarrow 17\%$ & 78.312                        &  0.057655 & \textbf{10.345 } min       & \textbf{7.6 } M                                       \\ \hline
\end{tabular}} \label{tabSWAG}
\vspace{-0.2cm}
\end{table}

\textbf{Natural Language Processing.} We perform lossless factorization of the Bert model on natural language processing tasks. And decompose all transformer blocks in the Bert model~\cite{devlin2018bert} that meet the algorithm requirements. The performance error ranges given in the table are within a reasonable range. Tables \ref{tabNLP} and \ref{tabSWAG} show the performance of the algorithm on SQuAD, MNLI, SWAG, and CoNLL-2003. Our algorithm outperforms existing decomposition methods in accuracy and loss and is comparable to the original model.
Existing decomposition methods require a large amount of data for fine-tuning and have a long running time. Our algorithm significantly reduces running time while maintaining a lossless compression model.
Our algorithm is a best-effort effort. In the SQuAD and MNLI datasets, the Bert model cannot satisfy the constraints of the lossless factorization optimization algorithm, so lossless factorization cannot be performed. However, when the loss is almost unchanged, our compact optimization algorithm can still compress the original model and maintain an accuracy similar to the original model. In the SWAG dataset, our algorithm can compress according to different needs and ensure that the loss is almost unchanged. In terms of parameter size, our algorithm can reduce the parameters of the Bert model by 17\% and ensure no loss.

\textbf{Algorithm Speed and Time.} The two algorithms proposed are best effort. Therefore, both algorithms are faster. After averaging multiple rounds of experiments, the running time of our algorithm is less than 10 minutes. At the same time, our algorithm does not require any fine-tuning. In terms of inference speed, our algorithm can achieve a 2\%-10\% reduction in inference time on image classification or language processing tasks while ensuring lossless.

\begin{table}[!ht]
\vspace{-0.2cm}
    \centering 
    \renewcommand{\arraystretch}{1}
    \setlength{\tabcolsep}{13pt}
    \caption{Ablation of gradients in lossless factorization.}
    \vspace{-0.2cm}
    \scalebox{0.8}{
    \begin{tabular}{c|c|c}
    \hline
        \textbf{Ablation} & \textbf{Accuracy} & \textbf{Loss} \\ \hline
        DenseNet161 & \textbf{77.138}  & 0.943676 \\ \hline
        DenseNet161-w/o gradient & 77.014  & 0.949644 \\ \hline
        DenseNet161-w/ gradient & 77.036  & \textbf{0.909673} \\ \hline
        DenseNet169 & \textbf{75.600}  & 0.997792 \\ \hline
        DenseNet169-w/o gradient & 75.442  & 0.997122 \\ \hline
        DenseNet169-w/ gradient & 75.446  & \textbf{0.971887} \\ \hline
        BERT\_base & \textbf{79.111}  & 0.057902 \\ \hline
        BERT\_base-w/o gradient & 78.421  & 0.058137 \\ \hline
        BERT\_base-w/ gradient & 78.566  & \textbf{0.056642} \\ \hline
    \end{tabular}} \label{tab6}
    \vspace{-0.4cm}
\end{table}

\subsection{Ablation}
\textbf{Optimization Objective.} In Table \ref{tab6}, when the algorithm optimizes for objectives that do not contain gradients, the loss increases. This illustrates the fact that approximating the weights through the original optimization objective increases the loss. The lossless factorization algorithm uses differential analysis to establish relationship between model performance loss and traditional factorization optimization and identifies a new optimization objective. It demonstrates the effectiveness of our optimization objective. 

\begin{table}[!ht]
\renewcommand{\arraystretch}{0.9}
\setlength{\tabcolsep}{30pt}
    \vspace{-0.1cm}
    \caption{Ablation of asymptotic upper bounds on noise $\epsilon$ introduced in low-rank factorizations in exponential form.}
    \vspace{-0.2cm}
    \scalebox{0.8}{
    \begin{tabular}{l|c|c}
    \hline
    \textbf{Layer} & \textbf{$\epsilon$} & \textbf{$\Delta$ $Loss$} \\ \hline
    \multirow{3}[0]{*}{Layer 1 } & O(1e-4) & O(9e-5) \\
          & O(5e-4) & O(3e-4) \\
          & O(1e-3) & O(2e-3) \\ \hline
    \multirow{3}[0]{*}{Layer 2} & O(1e-4) & O(6e-4) \\
          & O(5e-4) & O(6e-4) \\
          & O(1e-3) & O(1e-3) \\ \hline
    \multirow{5}[0]{*}{Layer 3} & O(1e-4) & O(9e-6) \\
          & O(5e-4) & O(7e-5) \\
          & O(1e-3) & O(3e-3) \\
          & O(1e-2) & O(5e-2) \\
          & O(1e-1) & O(7e-1) \\
           \hline
           
    \end{tabular}} \label{tababla}
\vspace{-0.2cm}
\end{table}


\textbf{Theory to Practice Mapping and Constraints.} We focus on whether the noise introduced by practical low-rank factorization satisfies our differential neighborhoods theory analysis. Our differential theory conditions that the neighborhoods need to be sufficiently small, and this corresponds to the performance implications of the actual factorization. We factorize the transformer layer. From Table \ref{tababla}, the factorization noise of the model's layers with different ranks leads to different upper bounds on the noise, and the variation in losses $\Delta$ $Loss$ due to this noise is very small. This satisfies our analysis of differential neighborhoods. When too much noise is introduced, the change in loss $\Delta$ $Loss$ is too large, which is also beyond the analytical scope of mathematical differentiation. Low rank introduces noise well beyond our lossless constraints. As shown in Figure \ref{fig_rank}, when the rank is too low, the noise introduced by factorization is too large, and the lossless restriction conditions in the optimization objective fail, causing the loss to rise beyond the original model. This makes our algorithm incapable of lossless factorization at current very low ranks.

\textbf{Higher-Order Term Effects and Weight Gradients.} In theoretical analysis, we found the first-order term is the main change causing model loss, while the influence of the second-order term is infinitesimal higher-order terms. We confirm this in our experiments, where the effect of the second-order term on the loss is generally less than $\mathcal{O}$(1e-5) when the same 1e-3 noise is introduced into the second-order term, while the effect of the first-order term on the loss is roughly 100 times more than the second-order term. The second-order term causes a change in accuracy in the range of about 0.001. So we mainly analyze the first-order term.
According to theoretical analysis, the ideal trained model weight gradient should be 0, but in practice, it is difficult to train a model so well that the gradient is 0. We experimentally found in DenseNet201, for example, about 99\% of the model gradient weight gradients are not 0, and most of them have to be less than 0.001. So our optimization objective's analysis of the gradient is effective in practical engineering.

\begin{table}[]  
    \centering  
    \renewcommand{\arraystretch}{1}  
    \setlength{\tabcolsep}{3pt}  
    \caption{Extensibility on Algorithms. In popular Vis-Transformer and Chat models, our algorithm has promising extensibility in different scenarios on Imagenet and MMLU~\cite{hendrycks2020measuring}.}  
    \vspace{-0.1cm}  
    \scalebox{0.83}{  
    \begin{tabular}{c|c|c|c|c|c}  
    \hline  
        \textbf{Vis-Model} & \textbf{Top-1}  & \textbf{Loss}  & \textbf{Chat Model} & \textbf{Accuracy}  & \textbf{SFT Loss} \\ \hline  
        Swin-T & 82.788 & 0.739740 & TinyLlama & 26.904 & 1.769011 \\ \hline  
        \rowcolor{gray!20} + Ours(L) & 82.790 & \textbf{0.736597} & + Ours(L) & \textbf{27.290} & \textbf{1.763389} \\ \hline  
        \rowcolor{yellow!20} + Ours(C) & \textbf{82.800} & 0.740220 & + Ours(C) & 26.884 & 1.769430 \\ \hline  
    \end{tabular}}  
    \label{tabex}  
    \vspace{-0.4cm}  
\end{table}

\subsection{Discussion}
\textbf{Optimization Objectives and Loss.}
Calculus leads to the analysis of lossless factorization. Eq.~\ref{eq:ctr_shale1} is one form of many solvable functions. We can further expand the optimization objective into the Eq.~\ref{eq:ctr_shale}
\begin{subequations}\label{eq:ctr_shale}
\vspace{-0.1cm}
\begin{align}
\operatorname*{min}G(l_{ij}, r_{ij})  \ \ \tag{\ref{eq:ctr_shale}}
\end{align} 
\vspace{-0.3cm}
\begin{alignat}{2}
\text{s.t.           } & U_{\delta^k}: \{\Vert w_{ij}-l_{ij}r_{ij} \Vert_F \}_{i,j} \leq\epsilon , \ \forall i,j   \\
    &0<k<\frac{NM}{N+M}
\end{alignat}
\end{subequations}
Where $G()$ represents the set of functions that satisfy the constraints. We solve the conditions for lossless factorization through total differentials, but the choice of factorization rank is discrete at this time, and we cannot solve it directly in finite space and time. To solve this problem, we need to construct a concrete function to describe $G()$. At this point, we utilize Eq.~\ref{eq7} to describe this function.
Our lossless algorithm is targeted at loss. Compared with other metrics, almost all tasks and models cannot be separated from analysis losses. Therefore, our algorithm compresses various models well to improve task performance.

\textbf{Structural Independence and Factorization Independence.} Our algorithm is derived from total differential analysis. For any model, we can write it in the form of a composite function and is therefore independent of any deep model structure. Our algorithm is also factorization-independent, that is, the algorithm does not rely on a specific low-rank factorization method. In Eq.~\ref{eq:ctr_shale1}, the factorization method can be replaced by other existing factorization methods. The product of noise and gradient depends on their respective directions. Therefore, our algorithm is universal when facing different replacement models or factorization methods. However, our experiments show that compared with other factorization methods, the original matrix factorization has the least parameters and running time.

\textbf{Extensibility and Future.} As shown in Table \ref{tabex}, our two algorithms are effectively extended to existing general vision and language models on Imagenet and MMLU, and remain lossless while compressing nearly 30\%. For large language models, we can still perform lossless decomposition. In future work, we will continue to expand the algorithm to large language models with larger parameter amounts.

\textbf{Limitation.} Due to varying layer sensitivities, lossless algorithm may find the same low-rank matrix with compression algorithm at the lowest loss. However when the rank is too low under extreme compression, factorization noise exceeds the differential domain, preventing lossless factorization. Therefore, we focus on achieving stable lossless factorization.

\vspace{-0.1cm}
\section{Conclusion}
This paper establishes the connection between the optimization of low-rank factorization and model optimization and converts the low-rank factorization problem into a numerical rank-defect problem under inequality constraints. We propose a lossless optimization algorithm and a compressed matrix optimization algorithm that does not require fine-tuning. Experiments show the effectiveness and generalization of our algorithm across various tasks.
{
    \small
    \bibliographystyle{ieeenat_fullname}
    \bibliography{main}
}


\end{document}